\documentclass[11pt,a4paper]{article}
\usepackage[hyperref]{acl2021}
\usepackage{times}
\usepackage{latexsym}

\usepackage{tikz}
\usepackage{overpic}
\usepackage{xcolor}
\usepackage{soul}
\usepackage{xstring}
\usepackage{inconsolata}
\usepackage{booktabs}
\usepackage{multirow}
\usepackage{enumitem}
\usepackage{relsize}

\usepackage[caption=false]{subfig}
\usepackage{soul}

\usepackage{amsmath,amsfonts,bm}

\def\eqref#1{equation~\ref{#1}}

\def\1{\bm{1}}

\DeclareMathAlphabet{\mathsfit}{\encodingdefault}{\sfdefault}{m}{sl}
\SetMathAlphabet{\mathsfit}{bold}{\encodingdefault}{\sfdefault}{bx}{n}

\usepackage{microtype}

\aclfinalcopy 
\def\aclpaperid{3604} 

\definecolor{rainbowred}{HTML}{FEE0E0}
\definecolor{actionbrown}{HTML}{DAC4AE}
\definecolor{boringgray}{HTML}{E9EAF0}
\definecolor{thormaterialcolor}{HTML}{ECF5E1}
\definecolor{thorstatecolor}{HTML}{03FCCA}

\newcommand{\button}[2]%
{\begin{tikzpicture}[baseline=(tempname.base)]
        \node[draw=gray, fill=#2, rounded corners=0.5pt, inner xsep=1pt, inner ysep=0.8pt, outer sep=0pt, minimum width=1.5em, minimum height=0em] (tempname) {\tt #1};
    \end{tikzpicture}}

\newcommand{\thorobject}[1]{\button{#1}{rainbowred}}
\newcommand{\blankthorobject}[1]{\button{#1}{boringgray}}

\newcommand{\thorfunction}[1]{\button{\StrSubstitute{#1}{ }{\detokenize{_}}}{boringgray}}
\newcommand{\thorstate}[1]{\button{#1}{thorstatecolor}}
\newcommand{\thoraction}[1]{\button{#1}{actionbrown}}

\newcommand{\mydots}{\ifmmode\mathinner{\ldotp\kern-0.2em\ldotp\kern-0.2em\ldotp}\else.\kern-0.13em.\kern-0.13em.\fi}
\newcommand{\mydotsc}{\ifmmode\mathinner{\kern-0.2em,\kern-0.1em\ldotp\kern-0.2em\ldotp\kern-0.2em\ldotp\kern-0.2em,}\else,\kern-0.1em.\kern-0.13em.\kern-0.13em.\kern-0.13em,\fi}

\newcommand\pigletfont[1]{{\usefont{T1}{curlzmt}{m}{n}#1}}

\newcommand{\modelname}{\pigletfont{PIGLeT}}
\newcommand{\modelnametitle}{{\huge\pigletfont{PIGLeT}}}

\newcommand{\modelnamelong}{\textbf{P}hysical \textbf{I}nteraction as \textbf{G}rounding for \textbf{L}anguag\textbf{e} \textbf{T}ransformers}
\newcommand{\dataname}{\pigletfont{PIGPeN}}
\newcommand{\datanamelong}{\textbf{P}hysical \textbf{I}nteraction \textbf{G}rounding \textbf{P}air\textbf{e}d with \textbf{N}atural Language}

\newcommand{\datanamenlu}{\pigletfont{PIGPeN}-NLU}
\newcommand{\datanamenlg}{\pigletfont{PIGPeN}-NLG}
\DeclareMathSymbol{\shortminus}{\mathbin}{AMSa}{"39}

\title{\modelnametitle:\\ Language Grounding Through Neuro-Symbolic Interaction in a 3D World}

\author{Rowan Zellers$^\spadesuit$ \: \: 
  Ari Holtzman$^{\spadesuit}$ \: \: 
  Matthew Peters$^{\heartsuit}$ \: \: \\ \bf
  Roozbeh Mottaghi$^{\heartsuit}$ \: \: 
  Aniruddha Kembhavi$^{\heartsuit}$ \: \: 
  Ali Farhadi$^{\spadesuit}$ \: \:
  Yejin Choi$^{\spadesuit\heartsuit}$\\
  $^\spadesuit$Paul G. Allen School of Computer Science \& Engineering, University of Washington \\
  $^\heartsuit$Allen Institute for Artificial Intelligence\\
  \vspace{1mm} \url{https://rowanzellers.com/piglet}
  }

\date{}

\begin{document}
\maketitle

\pagestyle{plain}
\thispagestyle{plain}

\begin{abstract}
We propose \modelname: a model that learns physical commonsense knowledge through interaction, and then uses this knowledge to ground language. We factorize \modelname~into a physical dynamics model, and a separate language model. Our dynamics model learns not just what objects \emph{are} but also what they \emph{do}: glass cups break when thrown, plastic ones don't. We then use it as the interface to our language model, giving us a unified model of linguistic form and grounded meaning. \modelname~can read a sentence, simulate neurally what might happen next, and then communicate that result through a literal symbolic representation, or natural language. 

Experimental results show that our model effectively learns world dynamics, along with how to communicate them. It is able to correctly forecast ``what happens next'' given an English sentence over 80\% of the time, outperforming a 100x larger, text-to-text approach by over 10\%. Likewise, its natural language summaries of physical interactions are also judged by humans as more accurate than LM alternatives. We present comprehensive analysis showing room for future work.

\end{abstract}

\section{Introduction}
\begin{figure}[t!]
\vspace{-3mm}
\centering\small
\includegraphics[width=\linewidth]{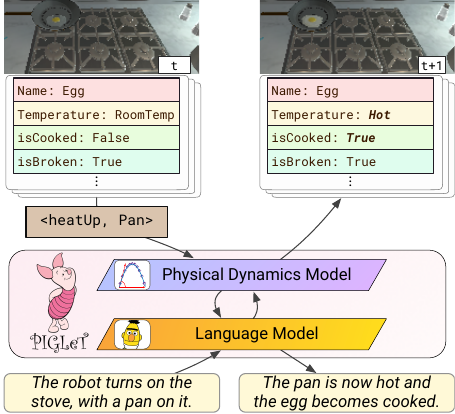}\vspace{-2mm}
  \caption{\modelname. Through physical interaction in a 3D world, we learn a model for what actions do to objects. We use our physical model as an interface for a language model, jointly modeling elements of language \emph{form} and \emph{meaning}. Given an action expressed symbolically or in English, \modelname~can simulate what might happen next, expressing it symbolically or in English.}
  \label{fig:teaser}
\end{figure}

As humans, our use of language is linked to the physical world. To process a sentence like  ``the robot turns on the stove, with a pan on it'' (Figure~\ref{fig:teaser}) we might imagine a physical \thorobject{Pan} object. This meaning representation in our heads can be seen as a part of our commonsense world knowledge, about what a \thorobject{Pan} is and does. We might reasonably predict that the \thorobject{Pan} will become \thorstate{Hot} -- and if there's an \thorobject{Egg} on it, it would become \thorstate{Cooked}.

As humans, we learn such a commonsense world model through interaction. Young children 
learn to reason physically about basic objects by manipulating them: observing the properties they have, and how they change if an action is applied on them \citep{smith2005development}. This process is hypothesized to be crucial to how children learn language: the names of these elementary objects become their first ``real words'' upon which other language is scaffolded \citep{yu2012embodied}. 



\begin{figure*}[t!]
\vspace{-3mm}
\centering\small
\includegraphics[width=\linewidth]{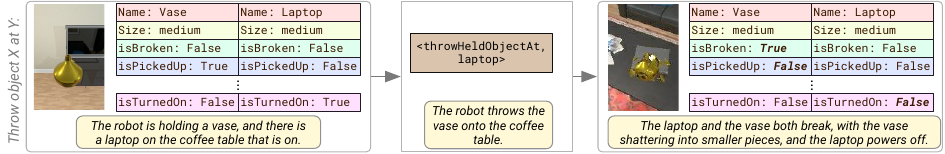}\vspace{-2mm}
\caption{\dataname, a setting for few-shot language-world grounding.
We collect data for 280k physical interactions in THOR, a 3D simulator with 20 actions and 125 object types, each with 42 attributes (e.g. {\footnotesize \texttt{isBroken}}). 
We annotate 2k interactions with English sentences describing the initial world state, the action, and the action result.}
  \label{fig:teaser2}
\end{figure*}

In contrast, the dominant paradigm today is to train large language or vision models on \emph{static data}, such as language and photos from the web. 
Yet such a setting is fundamentally limiting, as suggested empirically by psychologists' failed attempts to get kittens to learn passively \citep{held1963movement}. More recently, though large Transformers have made initial progress on benchmarks, they also have frequently revealed biases in those same datasets, suggesting they might not be solving underlying tasks \citep{zellers-etal-2019-hellaswag}.
This has been argued philosophically by a flurry of recent work arguing that no amount of language \emph{form} could ever specify language \emph{meaning} \citep{mcclelland2019extending,bender-koller-2020-climbing,bisk2020experience}; connecting back to the Symbol Grounding Problem of \citet{harnad1990symbol}.

In this paper, we investigate an alternate strategy for learning physical commonsense through interaction, and then transferring that into language. 
We introduce a model named~\modelname, short for \modelnamelong. We factorize an embodied agent into an explicit model of world dynamics, and a model of language form. We learn the dynamics model through \emph{interaction}. Given an action \thoraction{heatUp} applied to the \thorobject{Pan} in Figure~\ref{fig:teaser}, the model learns that the \thorobject{Egg} on the pan becomes \thorstate{Hot} and \thorstate{Cooked}, and that other attributes do not change. 

We integrate our dynamics model with a pretrained language model, giving us a joint model of linguistic \emph{form} and \emph{meaning}. The combined \modelname~can then reason about the physical dynamics implied by English sentences describing actions, predicting literally what might happen next. It can then communicate that result either symbolically or through natural language, generating a sentence like `The egg becomes hot and cooked.'' Our separation between physical dynamics and language allows the model to learn about physical commonsense from the physical world itself, while also avoiding recurring problems of artifacts and biases that arise when we try to model physical world understanding solely through language.


We study this through a new environment and evaluation setup called \dataname, short for \datanamelong. In \dataname, a model is given unlimited access to an environment for pretraining, but only 500 examples with paired English annotations. Models in our setup must additionally generalize to novel `unseen' objects for which we intentionally do not provide paired language-environment supervision.  We build this on top of the THOR environment \cite{kolve2017ai2}, a physics engine that enables agents to perform contextual interactions (Fig~\ref{fig:teaser2}) on everyday objects. 

Experiments confirm that \modelname~performs well at grounding language with meaning. Given a sentence describing an action, our model predicts the resulting object states correctly over 80\% of the time, outperforming even a 100x larger model (T5-11B)  by over 10\%. Likewise, its generated natural language is rated by humans as being more correct than equivalently-sized language models. Last, it can generalize in a `zero-shot' way to objects that it has never read about before in language.

In summary, we make three key contributions. \textbf{First}, we introduce \modelname, a model decoupling physical and linguistic reasoning. \textbf{Second}, we introduce \dataname, to learn and evaluate the transfer of physical knowledge to the world of language. \textbf{Third}, we perform experiments and analysis suggesting promising avenues for future work. 

\section{\dataname: A Resource for Neuro-Symbolic Language Grounding}
We introduce \dataname~as a setting for learning and evaluating physically grounded language understanding. An overview is shown in Figure~\ref{fig:teaser2}. The idea is that an agent gets access to an interactive 3D environment, where it can learn about the world through interaction -- for example, that objects such as a \thorobject{Vase} can become \thorstate{Broken} if thrown. The goal for a model is to learn natural language \emph{meaning} grounded in these interactions.


\textbf{Task definition.} Through interaction, an agent observes the interplay between objects $\boldsymbol{o} \in \mathcal{O}$ (represented by their attributes) and actions $\boldsymbol{a} \in \mathcal{A}$ through the following transition:
{\setlength{\abovedisplayskip}{3pt}
\setlength{\belowdisplayskip}{4pt}
\begin{equation}
\label{eqn:transition}
    \underbrace{\{\boldsymbol{o}_1, \ldots, \boldsymbol{o}_N\}}_{\vec{\boldsymbol{o}}, \textrm{ state pre-action}} \times \boldsymbol{a} \rightarrow \underbrace{\{\boldsymbol{o}'_1, \ldots, \boldsymbol{o}'_N\}}_{\vec{\boldsymbol{o}}', \textrm{ state post-action}}.
\end{equation}
}
Actions change the state of a subset of objects: turning on a \thorobject{Faucet} affects a nearby \thorobject{Sink}, but it will not change a \thorobject{Mirror} on the wall. 

To encourage learning from interaction, and not just language, an agent is given a small number of natural language annotations of transitions. We denote these sentences as $\boldsymbol{s}_{\vec{\boldsymbol{o}}}$, describing the state pre-action, $\boldsymbol{s}_{\boldsymbol{a}}$ the action, and $\boldsymbol{s}_{\vec{\boldsymbol{o}}'}$ the state post-action respectively. During evaluation, an agent will sometimes encounter new objects $\boldsymbol{o}$ that were not part of the paired training data.


We evaluate the model's transfer in two ways:
\begin{enumerate}[wide,labelwidth=!,listparindent=0pt, labelindent=-1pt,noitemsep,topsep=0pt,parsep=2pt,leftmargin =*,label=\textbf{\alph*}.]
    \item \datanamenlu. A model is given object states $\vec{\boldsymbol{o}}$, and an English sentence $\boldsymbol{s}_{\boldsymbol{a}}$ describing an action. It must predict the grounded object states $\vec{\boldsymbol{o}}'$ that result after the action is taken.
    \item \datanamenlg. A model is given object states $\vec{\boldsymbol{o}}$ and a literal action $\boldsymbol{a}$. It must generate a sentence $\boldsymbol{s}_{\vec{\boldsymbol{o}}'}$ describing the state post-action.
\end{enumerate}

We next describe our environment, feature representation, and language annotation process.

\subsection{Environment: THOR}
We use AI2-THOR as an environment for this task \cite{kolve2017ai2}. In THOR, a robotic agent can navigate around and perform rich contextual interactions with objects in a house. For instance, it can grab an \thorobject{Apple}, slice it, put it in a \thorobject{Fridge}, drop it, and so on. The state of the \thorobject{Apple}, such as whether it is sliced or cold, changes accordingly; this is not possible in many other environments.

In this work, we use the underlying THOR simulator as a proxy for grounded meaning. Within THOR, it can be seen as a `complete' meaning representation \cite{artzi2013semantic}, as it fully specifies the kind of grounding a model can expect in its perception within THOR. 

\textbf{Objects.} The underlying THOR representation of each object $\boldsymbol{o}$ is in terms of 42 attributes; we provide a list in Appendix~\ref{sec:appendix-thor-attributes}. We treat these attributes as words specific to an attribute-level dictionary; for example, the temperature \thorstate{Hot} is one of three possible values for an object's temperature; the others being \thorstate{Cold} and \thorstate{RoomTemp}.

\textbf{Actions.} An action $\boldsymbol{a}$ in THOR is a function that takes up to two objects as arguments. Actions are highly contextual, affecting not only the arguments but potentially other objects in the scene (Figure~\ref{fig:teaser2}). 
We also treat action names as words in a dictionary.

\textbf{Filtering out background objects.} 
Most actions change the state of only a few objects, yet there can be many objects in a scene. We keep annotation and computation tractable by having models predict (and humans annotate) possible changes of at most two key objects in the scene. As knowing when an object \emph{doesn't} change is also important, we include non-changing objects if fewer than two change.

\textbf{Exploration.} Any way of exploring the environment is valid for our task, however, we found that exploring \emph{intentionally} was needed to yield good coverage of interesting states. Similar to prior work for instruction following \cite{shridhar2020alfred}, we designed an oracle to collect diverse and interesting trajectories $\{\vec{\boldsymbol{o}}, \boldsymbol{a}, \vec{\boldsymbol{o}}'\}$. Our oracle randomly selects one of ten high level tasks, see Appendix~\ref{sec:appendix-thor-attributes} for the list. These in turn require randomly choosing objects in the scene; e.g. a \thorobject{Vase} and a \thorobject{Laptop} in Figure~\ref{fig:teaser2}. We randomize the manner in which the oracle performs the task to discover diverse situations.

In total, we sampled 20k trajectories. From these we extracted 280k transitions (Eqn~\ref{eqn:transition}'s) where at least one object changes state, for training. 

\subsection{Annotating Interactions with Language}


\subsubsection{Data Selection for Annotation}
We select 2k action state-changes from trajectories held out from the training set. We select them while also balancing the distribution of action types to ensure broad coverage in the final dataset. 
We are also interested in a model's ability to generalize to new object categories -- beyond what it has read about, or observed in a training set. We thus select 30 objects to be ``unseen,'' and exclude these from paired environment-language training data. We sample 500 state transitions, containing only ``seen'' objects to be the training set; we use 500 for validation and 1000 for testing.

\subsubsection{Natural Language Annotation}
Workers on Mechanical Turk were shown an environment in THOR \emph{before} and \emph{after} a given action $\boldsymbol{a}$. Each view contains the THOR attributes of the two key objects. Workers then wrote three English sentences, corresponding to $\boldsymbol{s}_{\vec{\boldsymbol{o}}}$, $\boldsymbol{s}_{\boldsymbol{a}}$, and $\boldsymbol{s}_{\vec{\boldsymbol{o}}'}$ respectively. Workers were instructed to write at a particular level of detail: enough so that a reader could infer ``what happens next'' from $\boldsymbol{s}_{\vec{\boldsymbol{o}}}$ and $\boldsymbol{s}_{\boldsymbol{a}}$, yet without mentioning redundant attributes.
We provide more details in Appendix~\ref{sec:turkstuff}.

\begin{figure*}[t!]
\centering\small\vspace{-2mm}
\includegraphics[width=0.9\linewidth]{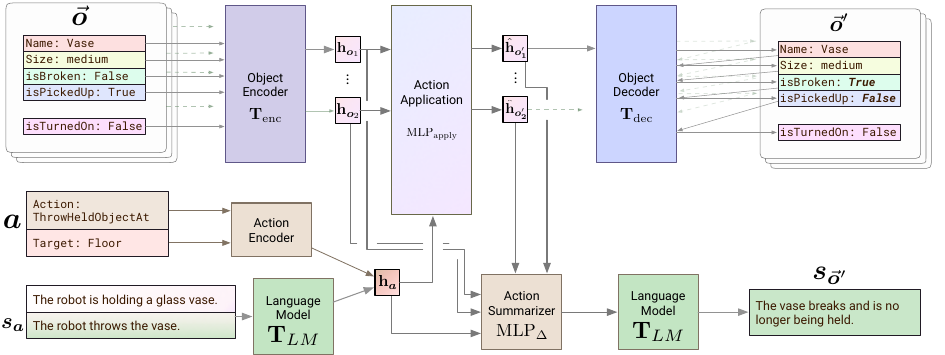}\vspace{-2mm}
  \caption{\modelname~architecture. We pretrain a model of physical world dynamics by learning to transform objects $\vec{\boldsymbol{o}}$ and actions $\boldsymbol{a}$ into new updated objects $\vec{\boldsymbol{o}}'$. Our underlying world dynamics model -- the encoder, the decoder, and the action application module, can augment a language model with grounded commonsense knowledge.}
  \label{fig:modelfig}
\end{figure*}

\section{Modeling \modelname}
In this section, we describe our \modelname~model. \textbf{First}, we learn a neural physical dynamics model from interactions, and \textbf{second}, integrate with a pretrained model of language form. 

\subsection{Modeling Physical Dynamics}
We take a neural, auto-encoder style approach to model world dynamics. An object $\boldsymbol{o}$ gets encoded as a vector $\mathbf{h}_{\boldsymbol{o}} \in \mathbb{R}^{d_{\textrm{o}}}$. The model likewise encodes an action $\boldsymbol{a}$ as a vector $\mathbf{h}_{\boldsymbol{a}} \in \mathbb{R}^{d_{\textrm{a}}}$, using it to manipulate the hidden states of all objects. The model can then decode any object hidden representation back into a symbolic form. 


\subsubsection{Object Encoder and Decoder}
\label{sssec:objencoder}
We use a Transformer \cite{vaswani2017attention} to encode objects into vectors $\mathbf{o} \in \mathbb{R}^{d_{\textrm{o}}}$, and then another to decode from this representation.

\textbf{Encoder.} Objects $\boldsymbol{o}$ are provided to the encoder as a set of attributes, with categories $c_1\mydotsc c_{n}$. Each attribute $c$ has its own vocabulary and embedding $\mathbf{E}_{c}$. For each object $\boldsymbol{o}$, we first embed all the attributes separately and feed the result into a Transformer encoder $T_{enc}$. This gives us (with position embeddings omitted for clarity):
{\setlength{\abovedisplayskip}{3pt}
\setlength{\belowdisplayskip}{4pt}
\begin{equation}
    \mathbf{h}_{\boldsymbol{o}} = \mathbf{T}_{\textrm{enc}}\Big(\mathbf{E}_{1}(o_{1}), \dotsc, \mathbf{E}_{c_n}(o_{c_n})\Big)
\end{equation}
}
\textbf{Decoder.} We can then convert back into the original symbolic representation through a left-to-right Transformer decoder, which predicts attributes one-by-one from $c_1$ to $c_n$. This captures the inherent correlation between attributes, while making no independence assumptions, we discuss our ordering in Appendix~\ref{ssec:appendix-att-ordering}.
The probability of predicting the next attribute $o_{c_{i+1}}$ is then given by:
{\setlength{\abovedisplayskip}{3pt}
\setlength{\belowdisplayskip}{4pt}
\begin{equation}
\label{eqn:predictprob}
\kern-0.8em p(o_{c_{i+1}}\kern-0.1em|\mathbf{h}_{\boldsymbol{o}}, \boldsymbol{o}_{:c_{i}}\kern-0.1em){=}  \mathbf{T}_{\textrm{dec}}\kern-0.1em\Big(\kern-0.1em\mathbf{h}_{\boldsymbol{o}},\kern-0.15em\mathbf{E}_{1}\kern-0.2em(o_{1}\kern-0.1em)\mydotsc \mathbf{E}_{c_i}\kern-0.1em(o_{c_i}\kern-0.1em)\kern-0.1em\Big)\kern-0.5em
\end{equation}
}
\subsubsection{Modeling actions as functions}
We treat actions $\boldsymbol{a}$ as functions that transform the state of all objects in the scene. Actions in our environment take at most two arguments, so we embed the action $\boldsymbol{a}$ and the names of its arguments, concatenate them, and pass the result through a multilayer perceptron; yielding a vector representation $\mathbf{h}_{\boldsymbol{a}}$. 

\textbf{Applying Actions.} We use the encoded action $\mathbf{h}_{\boldsymbol{a}}$ to transform all objects in the scene, obtaining updated representations $\hat{\mathbf{h}}_{\boldsymbol{o}'}$ for each one. We take a \emph{global} approach, jointly transforming all objects. This takes into account that interactions are contextual: turning on a \thorobject{Faucet} might fill up a \thorobject{Cup} if and only if there is one beneath it.

Letting the observed objects in the interaction be $\boldsymbol{o}_1$ and $\boldsymbol{o}_2$, with encodings $\mathbf{h}_{\boldsymbol{o_1}}$ and $\mathbf{h}_{\boldsymbol{o_2}}$ respectively, we model the transformation via the following multilayer perceptron:
{\setlength{\abovedisplayskip}{3pt}
\setlength{\belowdisplayskip}{4pt}
\begin{equation}
\label{eqn:mlp_apply}
[\hat{\mathbf{h}}_{\boldsymbol{o_1}'}, \hat{\mathbf{h}}_{\boldsymbol{o_2}'}] = \textrm{MLP}_{\textrm{apply}}\Big(\big[\mathbf{h}_{\boldsymbol{a}}, \mathbf{h}_{\boldsymbol{o_1}}, \mathbf{h}_{\boldsymbol{o}_2}\big]\Big).
\end{equation}
}
The result can be decoded into symbolic form using the object decoder (Equation~\ref{eqn:predictprob}).

\subsubsection{Loss function and training}
We train our dynamics model on ($\vec{\boldsymbol{o}}$,$\boldsymbol{a}$,$\vec{\boldsymbol{o}}'$) transitions. The model primarily learns by running $\vec{\boldsymbol{o}}$,$\boldsymbol{a}$ through the model, predicting the updated output state $\hat{\mathbf{h}}_{\boldsymbol{o}'}$, and minimizing the cross-entropy of generating attributes of the real changed object $\vec{\boldsymbol{o}}'$. We also regularize the model by encoding objects $\vec{\boldsymbol{o}}, \vec{\boldsymbol{o}}'$ and having the model learn to reconstruct them. We weight all these cross-entropy losses equally. 
We discuss our architecture in Appendix~\ref{ssec:appendix-physical-dynamics-model}; it uses 3-layer Transformers, totalling 17M parameters. 



\subsection{Language Grounding} 
After pretraining our physical dynamics model, we integrate it with a Transformer Language Model (LM). In our framework, the role of the LM will be to both encode natural language sentences of actions into a hidden state approximating $\mathbf{h}_{\boldsymbol{a}}$, as well as summarizing the result of an interaction ($\vec{\boldsymbol{o}}$,$\boldsymbol{a}$,$\vec{\boldsymbol{o}}'$) in natural language.

\textbf{Choice of LM.} Our framework is compatible with any language model. However, to explore the impact of pretraining data on grounding later in this paper, we pretrain our own with an identical architecture to the smallest GPT2 (\citet{radford2019gpttwo}; 117M). To handle both classification and generation well, we mask only part of the attention weights out, allowing the model to encode a ``prefix'' bidirectionally; it generates subsequent tokens left-to-right \cite{dong2019unified}. We pretrain the model on Wikipedia and books; details in Appendix~\ref{sec:mylm}.

We next discuss architectural details of performing the language transfer, along with optimization.

\subsubsection{Transfer Architecture}
\textbf{English actions to vector form.} Given a natural language description $\boldsymbol{s}_{\boldsymbol{a}}$ of an action $\boldsymbol{a}$, like ``The robot throws the vase,'' for \datanamenlu,  our model will learn to parse this sentence into a neural representation $\mathbf{h}_{\boldsymbol{a}}$, so the dynamics model can simulate the result. We do this by encoding $\boldsymbol{s}_{\boldsymbol{a}}$ through our language model, $\mathbf{T}_{LM}$, with a learned linear transformation over the resulting (bidirectional) encoding. The resulting vector $\mathbf{h}_{\boldsymbol{s}_{\boldsymbol{a}}}$ can then be used by Equation~\ref{eqn:mlp_apply}.

\textbf{Summarizing the result of an action.} For \datanamenlg, our model simulates the result of an action $\boldsymbol{a}$ neurally, resulting in a predicted hidden state $\hat{\mathbf{h}}_{\boldsymbol{o}}$ for each object in the scene $\boldsymbol{o}$. To write an English summary describing ``what changed,'' we first learn a lightweight fused representation of the transition, aggregating the initial and final states, along with the action, through a multilayer perceptron. For each object $\boldsymbol{o}_i$ we have:

\begin{equation}
\label{eqn:textsum}
\mathbf{h}_{\Delta\boldsymbol{o}_i} = \textrm{MLP}_{\Delta}([\mathbf{h}_{\boldsymbol{o}_i}, \hat{\mathbf{h}}_{\boldsymbol{o_i}'}, \mathbf{h}_{\boldsymbol{a}}]).
\end{equation}

We then use the sequence $[\mathbf{h}_{\Delta\boldsymbol{o}_1}, \mathbf{h}_{\Delta\boldsymbol{o}_2}]$ as bidirectional context for our our LM to decode from. Additionally, since our test set includes novel objects not seen in training, we provide the names of the objects as additional context for the LM generator (e.g. `Vase, Laptop'); this allows a LM to copy those names over rather than hallucinate wrong ones. Importantly we only provide the surface-form names, \textbf{not} underlying information about these objects or their usage as with few-shot scenarios in the recent GPT-3 experiments \cite{brown2020language} -- necessitating that \modelname~learns what these names \emph{mean} through interaction.

\subsubsection{Loss functions and training.}
Modeling text generation allows us to incorporate a new loss function, that of minimizing the log-likelihood of generating each $\boldsymbol{s}_{\vec{\boldsymbol{o}}'}$ given previous words and the result of Equation~\ref{eqn:textsum}:
\begin{equation}
\label{eqn:genloss}
p(s^{\textrm{post}}_{i+1} | \boldsymbol{s}_{\vec{\boldsymbol{o}'},1:i}) = \mathbf{T}_{\textrm{LM}}(\mathbf{h}_{\Delta\boldsymbol{o}_1}, \mathbf{h}_{\Delta\boldsymbol{o}_2},\boldsymbol{s}_{\vec{\boldsymbol{o}'},1:i}).
\end{equation}
We do the same for the object states $\boldsymbol{s}_{\vec{\boldsymbol{o}}}$ pre-action, using $\mathbf{h}_{\boldsymbol{o}_i}$ as the corresponding hidden states. 

For \datanamenlu, where no generation is needed, optimizing Equation~\ref{eqn:textsum} is not strictly necessary. However, as we will show later, it helps provide additional signal to the model, improving overall accuracy by several percentage points.







\section{Experiments}
\vspace{-1mm}
\begin{table*}[t]
\defcitealias{gupta2019effective}{G\&D2019}
\defcitealias{alberti2019fusion}{Alberti et al.2019}
\centering\footnotesize
\resizebox{1.00\linewidth}{!}{
\subfloat{
\begin{tabular}{@{}p{0.4cm} @{\hspace{0.1cm}} p{4.9cm} c @{\hspace{0.2cm}} ccc @{}} \toprule
& \multirow{3}{*}{Model} & \multicolumn{4}{@{}c@{\hspace{2em}}}{Accuracy (\%)}    \\ \cmidrule{3-6}
& & \multirow{2}{*}{Val} & \multicolumn{3}{@{}c@{\hspace{2em}}}{Test} \\ \cmidrule{4-6} \cmidrule{4-6}
& &  & Overall & Seen & Unseen \\ \cmidrule{2-6}
&                         No Change                          &      27.4&      25.5&      29.9&      24.0\\ \cmidrule{2-6}
\multirow{6}{*}{\rotatebox[origin=c]{90}{{\footnotesize text-to-text}}}                                                
&                    GPT3-175B \cite{brown2020language}&      23.8&      22.4&      22.4&      21.4\\
&                                                      T5-11B \cite{raffel2019t5}&      68.5&      64.2&      79.5&      59.1\\
&                                                       T5-3B&  66.6&      63.3&      77.1&      58.7\\
&                                                       T5-Large & 56.5&      54.1&      69.2&      49.1\\
&                                                     T5-Base&      56.0&      53.9&      69.2&      48.8\\
&                                                    T5-Small&      39.9&      36.2&      57.0&      38.0\\ \cmidrule{2-6}
\multirow{4}{*}{\rotatebox[origin=c]{90}{{\footnotesize BERT style}}} 
&\citetalias{alberti2019fusion}, Pretrained Dynamics&      61.3&      53.9&      71.4&      48.1\\
&\citetalias{alberti2019fusion}&       \phantom{0}9.7&       \phantom{0}6.8&      16.2&       \phantom{0}3.7\\
& \citetalias{gupta2019effective}, Pretrained Dynamics&      43.8&      35.3&      60.9&      26.9\\
&\citetalias{gupta2019effective}&      15.1&      11.3&      23.1&       \phantom{0}7.3\\ \cmidrule{2-6}
&\modelname &      \textbf{81.8}&     \textbf{81.1}&      \textbf{83.8}&      \textbf{80.2}\\
\end{tabular}
}\hspace{0.3cm}
\subfloat{
\begin{tabular}{@{}ccccc @{}} \toprule
\multicolumn{5}{@{}c@{\hspace{1em}}}{Attribute-level accuracy (Test-Overall,\%)}\\ \cmidrule{1-5}
size&distance&mass&Temperature&isBroken\\ \cmidrule{1-1} \cmidrule(lr){2-2} \cmidrule(lr){3-3} \cmidrule(lr){4-4} \cmidrule(lr){5-5} 
{\footnotesize 8-way} & {\footnotesize 8-way} & {\footnotesize 8-way} & {\footnotesize 3-way} & {\footnotesize boolean}  \\ \cmidrule{1-5}
          83.2&          84.1&          96.3&          86.0&         94.8\\ \cmidrule{1-5}
          73.7&          77.0&          89.5&          84.2&           94.7\\
          83.9&          88.9&          94.3&          95.4&         98.1\\
          81.6&          90.0&          94.0&          95.6&         98.4\\
          81.8&          84.6&          94.3&          96.3&       95.8\\
          81.1&          87.5&          93.6&          96.1&          96.5\\
          82.2&          84.9&          93.8&          89.6&            93.5\\ \cmidrule{1-5}
          87.7&          87.6&          97.5&             93.4&          97.5\\
          53.4&          43.6&          84.0&          88.1&           95.1\\
          83.0&          86.9&          94.0&          93.7&              97.4\\
          68.6&          47.3&          82.2&          88.3&          95.8\\ \cmidrule{1-5}
          \textbf{92.3}&          \textbf{91.9}&          \textbf{99.2}&          \textbf{99.8}&            \textbf{99.0}\\
\end{tabular}
}
}  
\vspace{-1mm}
\caption{\textbf{Overall results}. \textbf{Left}: we show the model accuracies at predicting all attributes of an object correctly. We compare \modelname~with `text-to-text' approaches that represent the object states as a string, along with BERT-style approaches with additional machinery to encode inputs or decode outputs. \modelname~outperforms a T5 model 100x its size (11B params) and shows gains over the BERT-style models that also model action dynamics through a language transformer. \textbf{Right}: we show several attribute-level accuracies, along with the number of categories per attribute; \modelname~outperforms baselines by over 4 points for some attributes such as size and distance.}
\label{tab:overallresults}
\vspace{-1.em}
\end{table*}

We test our model's ability to {encode language} into a grounded form (\datanamenlu), and decode that grounded form into language (\datanamenlg). 

\subsection{\datanamenlu~Results.}
We first evaluate models by their performance on \datanamenlu: given objects $\vec{\boldsymbol{o}}$, and a sentence $\boldsymbol{s}_{\boldsymbol{a}}$ describing an action, a model must predict the resulting state of objects $\vec{\boldsymbol{o}}'$. We primarily evaluate models by accuracy; scoring how many objects for which they got all attributes correct. We compare with the following strong baselines: 
\begin{enumerate}[wide, labelwidth=!,listparindent=0pt, labelindent=0pt,noitemsep,topsep=0pt,parsep=2pt,leftmargin =*,label=\textbf{\alph*}.]
    \item No Change: this baseline copies the initial state of all objects $\vec{\boldsymbol{o}}$ as the final state $\vec{\boldsymbol{o}}'$.
    \item GPT3-175B \cite{brown2020language}, a very large language model for `few-shot' learning using a prompt. For GPT3, and other text-to-text models, we encode and decode the symbolic object states in a JSON-style dictionary format, discussed in Appendix~\ref{ssec:appendix-json}.
    \item T5 \cite{raffel2019t5}. With this model, we use the same `text-to-text' format, however here we train it on the paired data from \dataname. We consider varying sizes of T5, from T5-Small -- the closest in size to \modelname, up until T5-11B, roughly 100x the size. 
    \item \cite{alberti2019fusion}-style. This paper originally proposed a model for VCR \cite{zellers2019vcr}, where grounded visual information is fed into a BERT model as tokens; the transformer performs the grounded reasoning. We adapt it for our task by using our base LM and feeding in object representations from our pretrained object encoder, also as tokens. Our object decoder predicts the object, given the LM's pooled hidden state. This is ``pretrained dynamics,'' we also consider a version without a randomly initialized dynamics model.
    \item \cite{gupta2019effective}-style. Thiso paper proposes using Transformers to model physical state, for tasks like entity tracking in recipes. Here, the authors propose decoding a physical state attribute (like \thorstate{isCooked}) by feeding the model a label-specific {\small \texttt{[CLS]}} token, and then mapping the result through a hidden layer. We do this and use a similar object encoder as our \cite{alberti2019fusion}-style baseline.
\end{enumerate}
We discuss hyperparameters in Appendix~\ref{ssec:appendix-hypers}. 

\textbf{Results.} From the results (Table~\ref{tab:overallresults}), we can draw several patterns. Our model, \modelname~performs best at getting all attributes correct; doing so over 80\% on both validation and test sets, even for novel objects not seen during training. The next closest model is T5-11B, which scores 68\% on validation. Though when evaluated on objects `seen' during training it gets 77\%, that number drops by over 18\% for unseen objects. On the other hand, \modelname~has a modest gap of 3\%. This suggests that our approach is particularly effective at connecting unpaired language and world representations. At the other extreme, GPT3 does poorly in its `few-shot' setting, suggesting that size is no replacement for grounded supervision. 

\modelname~also outperforms `BERT style' approaches that control for the same language model architecture, but perform the physical reasoning inside the language transformer rather than as a separate model. Performance drops when the physical decoder must be learned from few paired examples (as in \citet{gupta2019effective}); it drops even further when neither model is given access to our pretrained dynamics model, with both baselines then underperforming `No Change.' This suggests that our approach of having a physical reasoning model \emph{outside of} an LM is a good inductive bias.

\subsubsection{Ablation study}
\begin{table}[t!]
\centering\small
\vspace{-2mm}
\begin{tabular}{@{} @{\hspace{0.1cm}} p{5.1cm} @{\hspace{0.1cm}} r @{}}
\toprule
Model & Accuracy (val;\%) \\ \cmidrule{1-2}
\modelname, No Pretraining         &      10.4\\ \cmidrule{1-2}
\modelname, Non-global $\textrm{MLP}_{\textrm{apply}}$             &      72.0\\ \cmidrule{1-2}
\modelname, Global $\textrm{MLP}_{\textrm{apply}}$     &      78.5\\
\modelname, Global $\textrm{MLP}_{\textrm{apply}}$,  Gen. loss (\ref{eqn:genloss}) &      81.8\\ \cmidrule{1-2}
\modelname, Symbols Only (Upper Bound)       &      89.3\\
\end{tabular}\vspace{-2mm}

\caption{Ablation study on \datanamenlu's validation set. Our model improves 6\% by modeling global dynamics of all objects in the scene, versus applying actions to single objects in isolation. We improve another 3\% by adding an auxiliary generation loss.}
\label{tab:ablations}
\end{table}

In Table~\ref{tab:ablations}~we present an ablation study of \modelname's components. Of note, by using a global representation of objects in the world (Equation~\ref{eqn:mlp_apply}), we get over 6\% improvement over a local representation where objects are manipulated independently. We get another 3\% boost by adding a generation loss, suggesting that learning to generate summaries helps the model better connect the world to language. 
Last, we benchmark how much headroom there is on \datanamenlu~by evaluating model performance on a `symbols only' version of the task, where the symbolic action $\boldsymbol{a}$ is given explicitly to our dynamics model. This upper bound is roughly 7\% higher than \modelname, suggesting space for future work.

\subsection{\datanamenlg~Results}
Next, we turn to \datanamenlg: given objects $\vec{\boldsymbol{o}}$, and the literal next action $\boldsymbol{a}$, a model must generate a sentence $\boldsymbol{s}_{\vec{\boldsymbol{o}}'}$ describing what will change in the scene. We compare with the following baselines:
\begin{enumerate}[wide, labelwidth=!,listparindent=0pt, labelindent=0pt,noitemsep,topsep=0pt,parsep=2pt,leftmargin =*,label=\textbf{\alph*}.]
    \item T5. We use a T5 model that is given a JSON-style dictionary representation of both $\vec{\boldsymbol{o}}$ and $\boldsymbol{a}$, it is finetuned to generate summaries $\boldsymbol{s}_{\vec{\boldsymbol{o}}'}$.
    \item LM Baseline. We feed our LM hidden states $\mathbf{h}_{\boldsymbol{o}}$ from our pretrained encoder, along with its representation of $\boldsymbol{a}$. The key difference between it and \modelname~is that we do \textbf{not} allow it to simulate neurally what might happen next -- $\textrm{MLP}_{\textrm{apply}}$ is never used here.
\end{enumerate}
\textbf{Size matters.} Arguably the most important factor controlling the fluency of a language generator is its size \cite{kaplan2020scaling}. Since our LM could also be scaled up to arbitrary size, we control for size in our experiments and only consider models the size of GPT2-base (117M) or smaller; we thus compare against T5-small as T5-Base has 220M parameters. We discuss optimization and sampling hyperparameters in Appendix~\ref{ssec:appendix-hypers}. 

\textbf{Evaluation metrics.} We evaluate models over the validation and test sets. We consider three main evaluation metrics: BLEU \cite{papineni2002bleu} with two references, the recently proposed BERTScore \cite{Zhang2020BERTScoreET}, and conduct a human evaluation. Humans rate both the fluency of post-action text, as well as its faithfulness to true action result, on a scale from $-1$ to $1$.

\begin{table}[t!]\vspace{-2mm}

\centering\small
\begin{tabular}{@{}p{1.8cm}@{}c@{\hspace{0.8em}}c@{\hspace{0.5em}}||@{\hspace{0.5em}}c@{\hspace{0.8em}}c@{\hspace{0.5em}}||@{\hspace{0.5em}}c@{\hspace{0.7em}}r @{}}
\toprule
\multirow{2}{*}{Model} &\multicolumn{2}{@{\hspace{-0.75em}}c@{}}{BLEU} &\multicolumn{2}{@{\hspace{-0.75em}}c@{}}{BERTScore}& \multicolumn{2}{@{}c@{}}{Human {\tiny (test; $[\shortminus1, 1]$)} }\\ 
                       & Val & Test               & Val & Test                    & {\tiny Fluency} & {\tiny Faithfulness}\\ \cmidrule{1-7}
   T5&      46.6&      43.4&      82.2&      81.0&       0.82&       0.15\\
  LM Baseline&      44.6&      39.7&      81.6&      78.8&       0.91&       -0.13\\
      \modelname&      \textbf{49.0}&      \textbf{43.9}&      \textbf{83.6}&      \textbf{81.3}&       \textbf{0.92}&       \textbf{0.22}\\  \cmidrule{1-7}
     Human&      44.5&      45.6&      82.6&      83.3&       0.94 & 0.71        
\end{tabular}\vspace{-2mm}
\caption{Text generation results on \datanamenlg, showing models of roughly equivalent size (up to 117M parameters). Our \modelname~outperforms the LM baseline (using the same architecture but omitting the physical reasoning component) by 4 BLEU points, 2 BERTScore $F_1$ points, and 0.35 points in a human evaluation of language faithfulness to the actual scene.   }
\label{tab:resultsGEN}
\end{table}

\textbf{Results.} We show our results in Table~\ref{tab:resultsGEN}. Of note, \modelname~is competitive with T5 and significantly outperforms the pure LM baseline, which uses a pretrained encoder for object states, yet has the physical simulation piece $\textrm{MLP}_{\textrm{apply}}$ removed. This suggests that simulating world dynamics not only allows the model to predict what might happen next, it leads to more faithful generation as well.

\section{Analysis}
\begin{figure*}[t!]
\vspace{-2mm}
\centering\small
  \includegraphics[width=\linewidth]{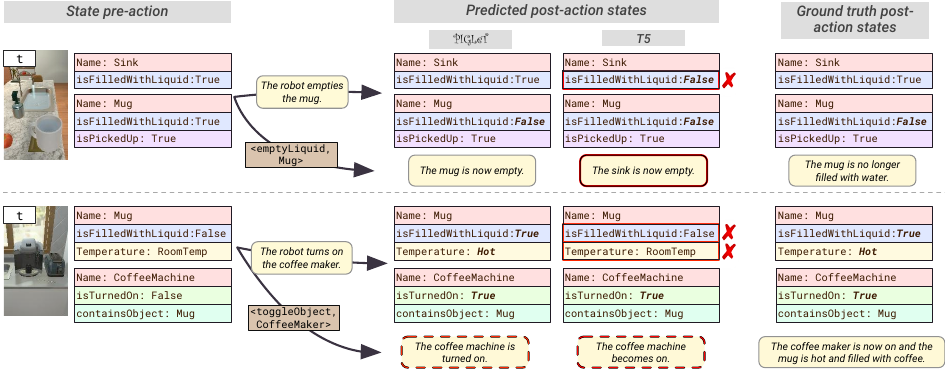}\vspace{-2mm}
  \captionof{figure}{Qualitative examples. Our model \modelname~reliably predicts what might happen next (like the \thorobject{Mug} becoming empty in Row 1), in a structured and explicit way. However, it often struggles at generating sentences for unseen objects like \thorobject{Mug} that are excluded from the training set. T5 struggles to predict these changes, for example, it seems to suggest that emptying the \thorobject{Mug} causes all containers in the scene to become empty.}
  \label{fig:qualex}
\end{figure*}

\subsection{Qualitative examples.}
We show two qualitative examples in Figure~\ref{fig:qualex}, covering both \datanamenlu~as well as \datanamenlg. In the first row, the robot empties a held \thorobject{Mug} that is filled with water. \modelname~gets the state, and generates a faithful sentence summarizing that the mug becomes empty. T5 struggles somewhat, emptying the water from both the \thorobject{Mug} and the (irrelevant) \thorobject{Sink}. It also generates text saying that the Sink becomes empty, instead of the Mug.

In the second row, \modelname~correctly predicts the next object states, but its generated text is incomplete -- it should also write that the mug becomes filled wtih Coffee. T5 makes the same mistake in generation, and it also underpredicts the state changes, omitting all changes to the \thorobject{Mug}.

We suspect that T5 struggles here in part because \thorobject{Mug} is an unseen object. T5 only experiences it through language-only pretraining, but this might not be enough for a fully grounded representation.

\subsection{Representing novel words}
\begin{figure}[t!]
\centering\small
  \includegraphics[width=\linewidth]{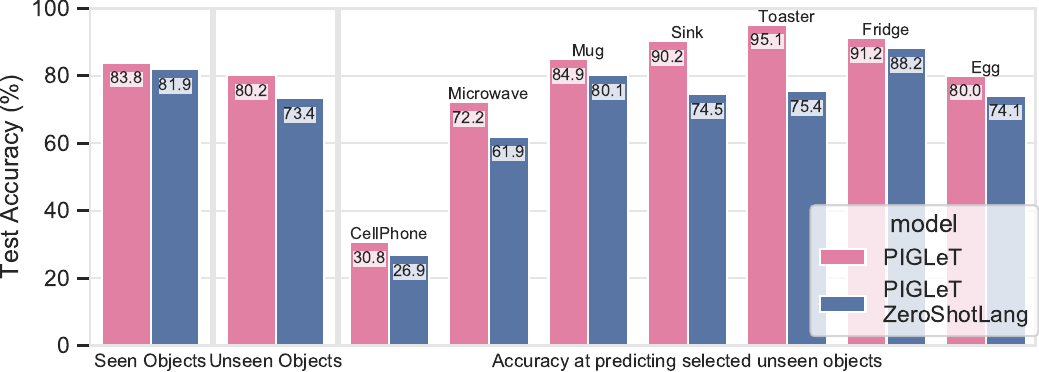}\vspace{-2mm}
  \captionof{figure}{\datanamenlu~performance of a \textbf{zero-shot} \modelname, that was pretrained on Books and Wikipedia without reading any words of our `unseen' objects like `mug.' It outperforms a much bigger T5-11B overall, though is in turn beaten by \modelname~on unseen objects like `Sink' and `Microwave.'}
  \label{fig:zslm}
\end{figure}

The language models that perform best today are trained on massive datasets of text. However, this has unintended consequences \cite{bender2021dangers} and it is unlike how children learn language, with children learning novel words from experience \cite{Carey1978AcquiringAS}. The large scale of our pretraining datasets might allow models to learn to perform physical-commonsense like tasks for wrong reasons, overfitting to surface patterns rather than learning meaningful grounding.

We investigate the extent of this by training a `zero-shot' version of our backbone LM on Wikipedia and books -- the only difference is that we explicitly \textbf{exclude} all mentioned sentences containing one of our ``unseen'' object categories. In this setting, not only must \modelname~learn to ground words like `mug,' it must do so without having seen the word `mug' during pretraining. This is significant because we count over 20k instances of `Mug' words (including morphology) in our dataset.

We show results in Figure~\ref{fig:zslm}. A version of \modelname~ with the zero-shot LM does surprisingly well -- achieving 80\% accuracy at predicting the state changes for ``Mug'' -- despite never having been pretrained on one before. This even outperforms T5 at the overall task. Nevertheless, \modelname~outperforms it by roughly 7\% at unseen objects, with notable gains of over 10\% on highly dynamic objects like \thorobject{Toaster}s and \thorobject{Sink}s.

\section{Related Work}
\textbf{Grounded commonsense reasoning}. In this work, we study language grounding and commonsense reasoning at the representation and concept level. The aim is to train models that learn to acquire concepts more like humans, rather than performing well on a downstream task that (for humans) requires commonsense reasoning. Thus, this work is somewhat different versus other 3D embodied tasks like QA \cite{Gordon_2018_CVPR, das2018embodied}, along with past work for measuring such grounded commonsense reasoning, like SWAG, HellaSWAG, and VCR \cite{zellers-etal-2018-swag,zellers-etal-2019-hellaswag, zellers2019vcr}. The knowledge covered is different, as it is self-contained within THOR. While VCR, for instance, includes lots of visual situations about what people are doing, this paper focuses on learning the physical properties of objects.

\textbf{Zero-shot generalization}. There has been a lot of past work involved with learning `zero-shot': often learning about the grounded world in language, and transferring that knowledge to vision. Techniques for this include looking at word embeddings \cite{frome2013devise} and dictionary definitions \cite{zellers2017zero}. In this work, we propose the inverse. This approach was used to learn better word embeddings \cite{gupta2019vico} or semantic tuples \cite{yatskar2016stating}, but we consider learning a component to be plugged into a deep Transformer language model.

Past work evaluating these types of zero-shot generalization have also looked into how well models can compose concepts in language together \cite{lake2018generalization,ruis2020benchmark}. Our work considers elements of compositionality through grounded transfer. For example, in \datanamenlg, models must generate sentences about the equivalent of dropping a `dax', despite never having seen one before. However, our work is also contextual, in that the outcome of `dropping a dax' might depend on external attributes (like how high we're dropping it from).


\textbf{Structured Models for Attributes and Objects}. The idea of modeling actions as functions that transform objects has been explored in the computer vision space \cite{Wang_Transformation}. Past work has also built formal structured models for connecting vision and language \cite{matuszek2012joint, krishnamurthy2013jointly}, we take a neural approach and connect today's best models of language \emph{form} to similarly neural models of a simulated environment. 

Past work has also looked into training neural models for a target domain -- similar to our factorized model for physical interaction. For example, \cite{leonandya2019fast} and \cite{gaddy2019pre} learn pretrained models for an instruction-following task in a blocks world, also using an autoencoder formulation. Our goal in this work is somewhat different: we are interested in learning physical reasoning about everyday objects, that might be discussed loosely in language (but with recurring issues of reporting bias \cite{gordon2013reporting}). We thus build a model that can be tied in with a pretrained language model, while also exhibiting generalization to new objects (that were not mentioned in language). We compare our model to today's largest language models that learn \emph{from text alone}, and find better performance despite having 100x fewer parameters.

\section{Conclusion}
In this paper, we presented an approach \modelname~for jointly modeling language form and meaning. We presented a testbed \dataname~for evaluating our model, which performs well at grounding language to the (simulated) world.

\section*{Acknowledgments}
We thank the reviewers for their helpful feedback, and the Mechanical Turk workers for doing a great job in annotating our data. Thanks also to Zak Stone and the Google Cloud TPU team for help with the computing infrastructure. This work was supported by the DARPA CwC program through ARO (W911NF-15-1-0543), the DARPA MCS program through NIWC Pacific (N66001-19-2-4031), and the Allen Institute for AI.

\bibliography{acl2021}
\bibliographystyle{acl_natbib}

\clearpage
\appendix


\section{Model implementation details and hyperparameters.}

We discuss the architectures and learning hyperparameters of our various models in the subsections below.

\subsection{Physical Dynamics Model}
\label{ssec:appendix-physical-dynamics-model}
We implemented our dynamics model with three Transformer layers for both the encoder and the decoder, and a hidden dimension of 256 for objects and actions. The resulting model has 17 million parameters. We pretrained the model for 20 epochs  over 280k state transitions, with a batch size of 1024. We use an Adam optimizer \cite{Kingma2014AdamAM} with a learning rate of 1e$^{-3}$.

\subsection{Ordering attributes in decoding.}
\label{ssec:appendix-att-ordering}
Recall that we use a left-to-right transformer to decode into an attribute representation, predicting attributes one-by-one from $c_1$ to $c_n$. Our model is agnostic to the actual order, as no matter what the order is, it still is modeling a decomposition of the joint probability of generating that object. However, we implemented this by using the name as the first attribute $c_1$ that is predicted, and ordered the rest in a descending way by vocabulary size so as to predict harder attributes first.

\subsection{Optimization Hyperparameters chosen}
\label{ssec:appendix-hypers}
We finetuned \modelname~for both tasks with an Adam optimizer \cite{Kingma2014AdamAM}. We did a small grid search for hyperparameter values, choosing the best learning rate $\{2e^{-5},1e^{-5},1e^{-6}\}$ by accuracy on the development set, and likewise the best batch size $16$ or $32$. We considered freezing the physical dynamics backbone as another hyperparameter. We found it slightly boosted performance on \datanamenlg~when we froze the physical dynamics backbone, but not so for \datanamenlu. We trained our model for 80 epochs on paired data.

We trained the baseline models with the same backbone in the same way, using similar hyperparameters. However, we found that after 80 epochs, the baseline models without pretrained dynamics failed to converge, so we finetuned them for 200 epochs total. 
For T5, we used similar identical hyperparameter ranges as the other models. However, because T5 uses a different optimizer (AdaFactor; \citet{shazeer2018adafactor}), which operates on a slightly different scale, we used a different set of learning rates. We chose the best one over $\{1e^{-4}, 2e^{-4}, 4e^{-4}\}$.

\textbf{Search.} Both of our tasks involve left-to-right decoding. We used argmax (greedy) search for \datanamenlu, finding that it worked well as a `closed-ended generation' style task. On the other hand, we used Nucleus Sampling for \datanamenlg~ as there are often several ways to communicate a state transition; here we set $p=0.8$.

\subsection{Encoding the input for text-to-text models}
\label{ssec:appendix-json}
Text-to-text models, needless to say, can only handle text. We encode the world states into a representation suitable for these models by formatting the object states as a JSON-style dictionary of keys and values. We had to make several modifications to the encoding however from a default JSON, because we handle a lot of attributes in this task, and JSON has quote characters `'` that take up a lot of space in a BPE encoding. We thus strip the quote characters and lowercase everything (with this also helping BPE-efficiency). We put parentheses around each object and give names to all `binned' attributes.

An example encoding might be:

{\footnotesize\tt Predict next object states: (objectname: bowl, parentreceptacles: cabinet, containedobjects: none, distance: 6 to 8 ft, mass: .5 to 1lb, size: medium, temp: roomtemp, breakable: true, cookable: false, dirtyable: true, broken: false, cooked: false, dirty: false, filledwithliquid: false, open: false, pickedup: false, sliced: false, toggled: false, usedup: false, moveable: false, openable: false, pickupable: true, receptacle: true, sliceable: false, toggleable: false, materials: glass) (objectname: egg, parentreceptacles: none, containedobjects: none, distance: 2 to 3ft, mass: .1 to .2lb, size: tiny, temp: cold, breakable: true, cookable: true, dirtyable: false, broken: false, cooked: false, dirty: false, filledwithliquid: false, open: false, pickedup: true, sliced: false, toggled: false, usedup: false, moveable: false, openable: false, pickupable: true, receptacle: false, sliceable: true, toggleable: false, materials: food) (action: throwobject10)}

We have models decode directly into this kind of format when predicting state changes. Though the T5 models usually get the format right, we often have to sanitize the text in order for it to be a valid object state in our framework. This is especially an issue with GPT3, since it is given limited supervision (we squeeze 3 examples into the 2048-BPE token context window) and often makes up new names and attributes. Thus, for each word not in an attribute's vocabulary, we use a Levenstein distance heuristic to match the an invalid choice with its closest (valid) option. If the model fails to generate anything for a certain attribute key -- for example if it does not include something like {\footnotesize\tt openable} somewhere, we copy the representation of the input object for that attribute, thereby making the default assumption that attributes do not change.

\section{All THOR attributes}
\label{sec:appendix-thor-attributes}
\begin{table}[t]
    \footnotesize\centering{\frenchspacing\renewcommand{\arraystretch}{1.0}
    \begin{tabular}{@{}p{3cm} @{\hspace{0.1cm}} c @{\hspace{0.1cm}} p{3cm}@{}}
        \toprule
Attribute Name & Vocab size & Values \\ \cmidrule{1-3}
objectName & 126 & One per object type, along with \thorobject{None} \\
parentReceptacles & 126 & One per object type, along with \thorobject{None} \\
receptacleObjectIds & 126 & One per object type, along with \thorobject{None} \\
mass & 8 & 8 bins \\
size & 8 & 8 bins \\
distance & 8 & 8 bins \\
ObjectTemperature & 3 & \thorstate{Hot}, \thorstate{Cold}, \thorstate{RoomTemp} \\
breakable & 2 &  \\
canBeUsedUp & 2 \\
canFillWithLiquid & 2 \\
cookable & 2 \\
dirtyable & 2 \\
isBroken & 2 \\
isCooked & 2 \\
isDirty & 2 \\
isFilledWithLiquid & 2 \\
isOpen & 2 \\
isPickedUp & 2 \\
isSliced & 2 \\
isToggled & 2 \\
isUsedUp & 2 \\
moveable & 2 \\
openable & 2 \\
pickupable & 2 \\
receptacle & 2 \\
sliceable & 1\\
toggleable& 2 \\
salientMaterials\_Ceramic & 2 \\
salientMaterials\_Fabric & 2 \\
salientMaterials\_Food & 2 \\
salientMaterials\_Glass & 2 \\
salientMaterials\_Leather & 2 \\
salientMaterials\_Metal & 2 \\
salientMaterials\_None & 2 \\
salientMaterials\_Organic & 2 \\
salientMaterials\_Paper & 2 \\
salientMaterials\_Plastic & 2 \\
salientMaterials\_Rubber & 2 \\
salientMaterials\_Soap & 2 \\
salientMaterials\_Sponge & 2 \\
salientMaterials\_Stone & 2 \\
salientMaterials\_Wax & 2 \\
salientMaterials\_Wood & 2 \\
         \bottomrule
    \end{tabular}\vspace{-1mm}}
    \caption{All attributes that we consider for this work in THOR. We list the attribute's name, the size of the attribute vocabulary, and the range of values the attribute can take on. For attributes like `mass', `size', and `distance', we note that the underlying simulator stores them as floats; we bin them to 8 values for this work. All the values for attributes with a vocabulary size of 2 are boolean.}
    \label{tab:thorattrs}
\end{table}

We list a table with all of the attributes we used for this work in Table~\ref{tab:thorattrs}.

\section{Turk Annotation Details}
\begin{figure}
  \centering\small
  \includegraphics[width=\linewidth]{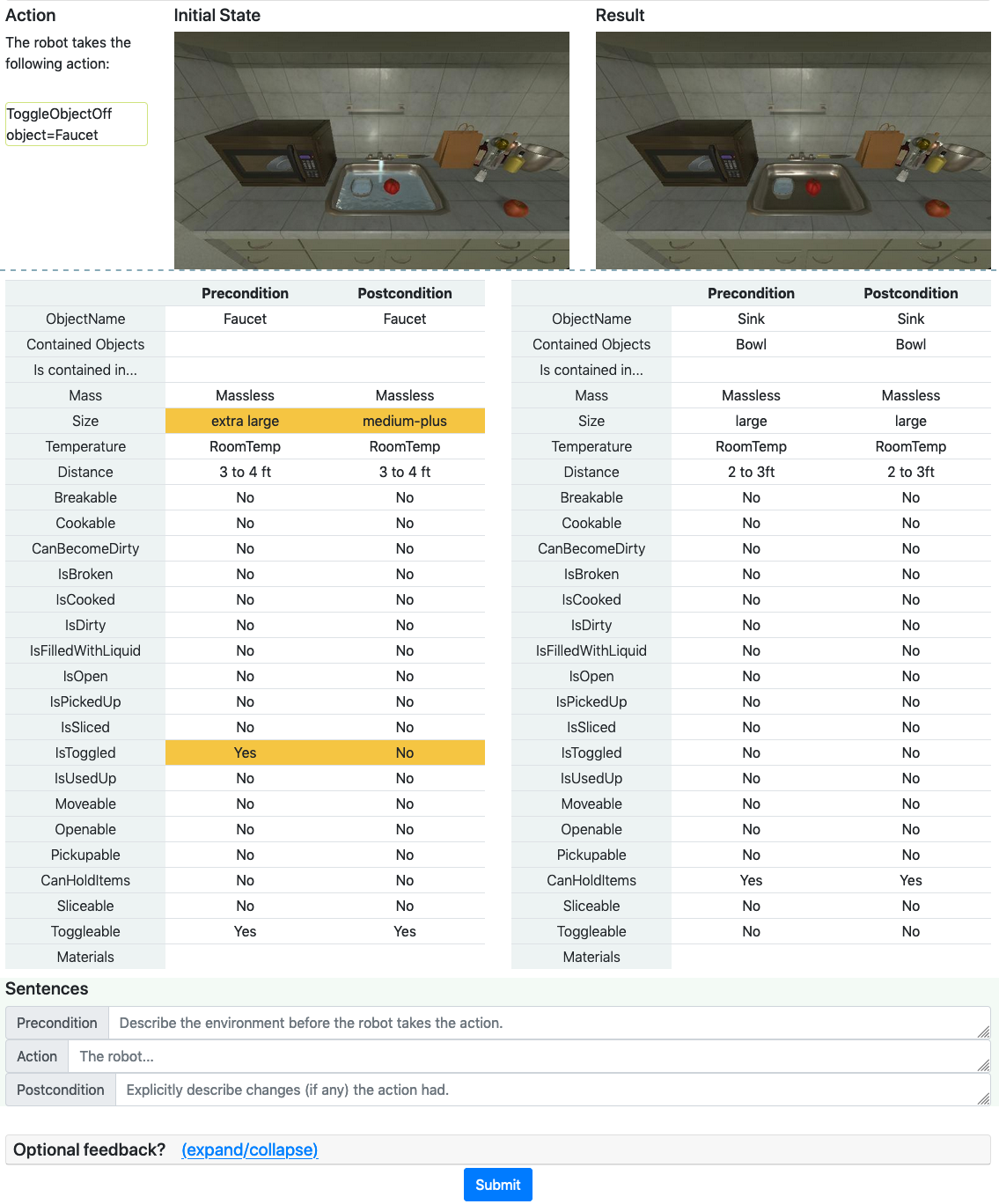}
  \caption{Our user interface for Mechanical Turk annotation.}
  \label{fig:turkui}
\end{figure}

We followed crowdsourcing best practices, such as using a qualification exam, giving feedback to workers, and paying workers well (above \$15 per hour). Each of our HITs required writing three sentences, and we paid Mechanical Turk workers 57 cents per HIT. We used three workers per example, allowing us to have multiple language references for evaluation. A screenshot of our user interface is shown in Figure~\ref{fig:turkui}.

\label{sec:turkstuff}

\begin{table}[t]
    \footnotesize\centering{\frenchspacing\renewcommand{\arraystretch}{1.3}
    \begin{tabular}{@{}p{2.2cm} @{\hspace{0.1cm}} p{5.6cm}@{}}
        \toprule
        Generator & Description \\[-0.05cm]
        \cmidrule{1-2}
        \thorfunction{put X in Y} & Samples an object \blankthorobject{X} from the scene, and a receptacle \blankthorobject{Y}. 
        Tries to put it in \blankthorobject{Y}. \\
        \thorfunction{throw X at Y} & Samples two objects \blankthorobject{X} and \blankthorobject{Y} from the scene. Picks up \blankthorobject{X}, moves to face \blankthorobject{Y}, and throws it forward with variable intensity. \\
        \thorfunction{toggle X} & Samples an object \blankthorobject{X}, and turns it on or off. \\
        \thorfunction{slice X} & Samples an object \blankthorobject{X} and a surface \blankthorobject{Y}. Picks up \blankthorobject{X}, places it on \blankthorobject{Y}, and cuts it. \\
        \thorfunction{dirty X} & Samples an object \blankthorobject{X}, and makes it dirty. \\
        \thorfunction{clean X} & Samples a dirty object \blankthorobject{X}. Finds a \thorobject{Sink} nearby a \thorobject{Faucet}, and places \blankthorobject{X} inside. Turns on/off the \thorobject{Faucet}, cleaning \blankthorobject{X}.\\
        \thorfunction{toast bread} & Finds some \thorobject{Bread}, slicing it if necessary, places it in a \thorobject{Toaster}, then turns it on. \\
        \thorfunction{brew coffee} & Picks up a \thorobject{Mug}, places it under a \thorobject{CoffeeMachine}, and turns the machine on. \\
        \thorfunction{fry X} & Picks up a food \blankthorobject{X}, slices it if necessary, and puts it in a \thorobject{Pot} or \thorobject{Pan}. Brings it to a \thorobject{StoveBurner} and turns the burner on.\\ 
        \thorfunction{microwave X} & Picks up an object \blankthorobject{X} and slices it if necessary. Places it in a \thorobject{Microwave}, closes it, and then turns it on. \\
        \thorfunction{fill X} & Picks up an object \blankthorobject{X} places it under a \thorobject{Faucet}. Turns on/off the \thorobject{Faucet}, then pours out the liquid.\\[-0.1cm]
         \bottomrule
    \end{tabular}\vspace{-1mm}}
    \caption{Trajectory generation functions that we used to sample `interesting' physical interactions, such as the effects that actions will have on objects, and which actions will succeed or not.}
    \label{tab:thortrajectoryfns}
\end{table}

\section{Our Pretrained Language Model}
\label{sec:mylm}

We use our own pretrained language model primarily because it allows us to investigate the impact of data on model performance. We trained a prefix-masked language model \cite{dong2019unified} on Wikipedia and Book data, mimicing the data used by the original BERT paper \cite{devlin2018bert}. We trained the model for 60000 iterations, at a batch size of 8192 sequences each of length 512. This corresponds to 50 epochs over the dataset. We masked inputs in the bidirectional prefix with SpanBERT masking \cite{joshi2020spanbert}. Since BERT-style `masked' out inputs are easier to predict than tokens generated left-to-right, we reduced the loss component of left-to-right generation by a factor of 20; roughly balancing the two loss components.

\begin{figure}
  \centering\small
  \includegraphics[width=\linewidth]{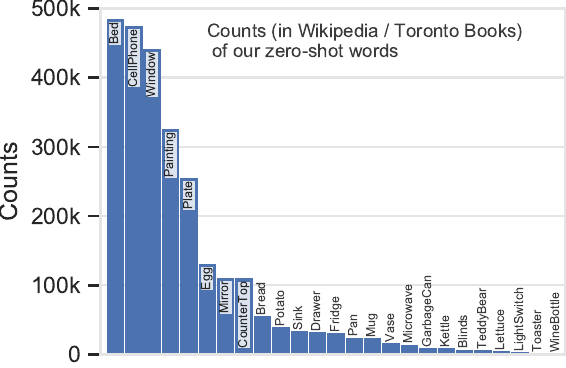}\vspace{5mm}
  \caption{Counts of zero-shot words that appear in BERT's training data (Wikipedia and Toronto Books). For example, in the 4 billion words BERT is trained on, it sees the word `Bed' almost 500k times. This might allow it to perform superficially well at answering questions about beds -- while not necessarily possessing deep physical knowledge about them.}
  \label{fig:conceptcounts}
\end{figure}

\end{document}